\documentclass{article}
\usepackage{spconf,amsmath,graphicx}
\usepackage{amsmath}
\usepackage{amssymb}
\usepackage{multirow}
\usepackage{makecell}
\usepackage{tabularx}
\usepackage{booktabs}
\usepackage{array}

\title{CountingFruit: Language-Guided 3D Fruit Counting with Semantic Gaussian Splatting}

\name{
\begin{tabular}{c}
Fengze Li$^{1,2,3\star}$, Yangle Liu$^{1,2,\star}$, Jieming Ma$^{2,\dagger}$, Hai-Ning Liang$^4$, \\
Yaochun Shen$^1$, Huangxiang Li$^3$, Zhijing Wu$^5$ \thanks{$\dagger$ Equally contributed. \quad $\star$ Corresponding author}
\end{tabular}
}
\address{\textit{University of Liverpool, Liverpool, UK$^1$},\\ 
\textit{Xi'an Jiaotong-Liverpool University, Suzhou, China$^2$},\\
\textit{Baidu AI Cloud Group (ACG), Beijing, China$^3$},\\
\textit{Hong Kong University of Science and Technology (Guangzhou), Guangzhou, China$^4$},\\
\textit{University of Cambridge, Cambridge, UK$^5$}}

\begin{document}
%\ninept
%
\maketitle
\begin{abstract}
Accurate 3D fruit counting in orchards is challenging due to heavy occlusion, semantic ambiguity between fruits and surrounding structures, and the high computational cost of volumetric reconstruction. Existing pipelines often rely on multi-view 2D segmentation and dense volumetric sampling, which lead to accumulated fusion errors and slow inference. We introduce FruitLangGS, a language-guided 3D fruit counting framework that reconstructs orchard-scale scenes using an adaptive-density Gaussian Splatting pipeline with radius-aware pruning and tile-based rasterization, enabling scalable 3D representation. During inference, compressed CLIP-aligned semantic vectors embedded in each Gaussian are filtered via a dual-threshold cosine similarity mechanism, retrieving Gaussians relevant to target prompts while suppressing common distractors (e.g., foliage), without requiring retraining or image-space masks. The selected Gaussians are then sampled into dense point clouds and clustered geometrically to estimate fruit instances, remaining robust under severe occlusion and viewpoint variation. Experiments on nine different orchard-scale datasets demonstrate that FruitLangGS consistently outperforms existing pipelines in instance counting recall, avoiding multi-view segmentation fusion errors and achieving up to 99.7\% recall on Pfuji-Size\_Orch2018 orchard dataset. Ablation studies further confirm that language-conditioned semantic embedding and dual-threshold prompt filtering are essential for suppressing distractors and improving counting accuracy under heavy occlusion. Beyond fruit counting, the same framework enables prompt-driven 3D semantic retrieval without retraining, highlighting the potential of language-guided 3D perception for scalable agricultural scene understanding.\end{abstract}
%
% \begin{keywords}
% Radiance Fields, Gaussian Splatting, Fruit Counting, Language-guided Filtering
% \end{keywords}
\section{Introduction}
The global transition toward smart agriculture has led to increasing demand for scalable, autonomous systems that can sense, reason, and act in dynamic outdoor environments~\cite{ss2024precision}. In this context, fruit counting plays a crucial role in yield prediction, robotic harvesting, and orchard monitoring, supporting critical decision-making in precision agriculture~\cite{krishnababu2024review}. Traditional manual methods are labor-intensive, error-prone, and lack scalability. As a result, vision-based techniques have gained traction across agricultural robotics, computer vision, and remote sensing communities~\cite{bhangar2023iot, neupane2024developing}.

Early approaches relied on 2D detection from RGB images using handcrafted features or deep object detectors~\cite{song2014automatic, rahnemoonfar2017deep}, which were often hampered by occlusions, varying lighting conditions, and inconsistent viewpoints~\cite{farjon2024agrocounters}. Recent works have leveraged multi-view geometry, introducing 3D-aware counting pipelines. A representative example is the use of Structure-from-Motion (SfM) to obtain camera poses and fuse 2D detections into 3D space~\cite{liu2018robust}, enabling more accurate fruit localization and preventing double counting. While effective, these pipelines typically require instance tracking or segmentation in every frame, limiting real-time applicability and robustness across diverse orchard scenes.

With the advent of neural scene representations, 3D reconstruction and semantic understanding have increasingly shifted toward learned volumetric representations. FruitNeRF~\cite{meyer2024fruitnerf} and FruitNeRF++~\cite{meyer2025fruitnerf++}, for instance, pioneered the use of semantic neural radiance fields for fruit counting, integrating RGB and segmentation masks into an implicit 3D representation. However, NeRF-based methods suffer from high computational cost and long rendering times, especially in agricultural environments where real-time feedback is essential for robotic deployment. Furthermore, semantic supervision in such pipelines typically relies on frame-level binary masks, requiring model-specific training and lacking open-vocabulary generalization.

\begin{figure*}[t]
    \centering
    \includegraphics[width=\textwidth]{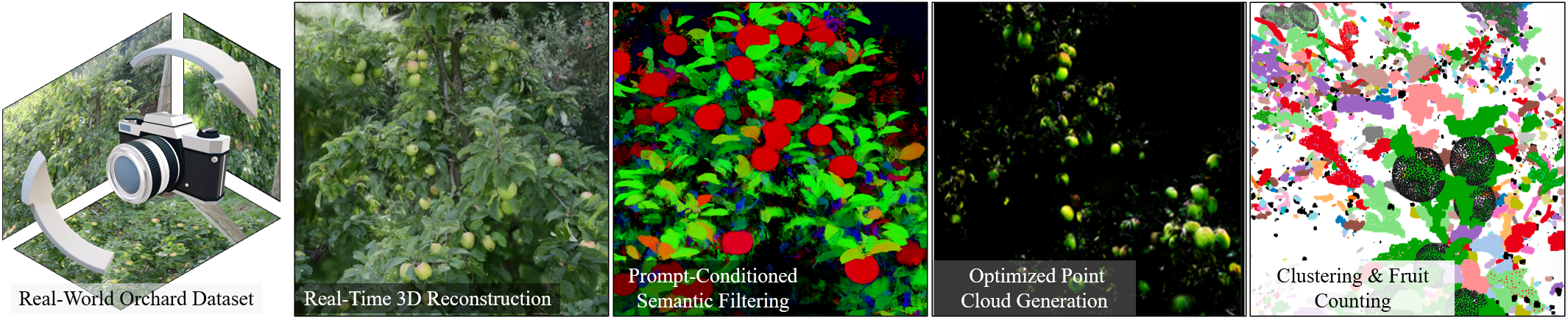}
    \caption{Overview of the FruitLangGS pipeline.}
  \label{fig:teaser}
\end{figure*}

To address the practical demands of agricultural robotics, an effective fruit counting system must not only localize fruits accurately under occlusion and semantic ambiguity, but also provide timely feedback for downstream tasks such as robot path planning, yield estimation, and harvest coordination. In this context, fast 3D rendering facilitates real-time perception and decision-making onboard field robots, while semantic controllability enables farmers or models to flexibly query different types of fruits without retraining for each crop or scene. 

We argue that a promising direction is to model the scene geometry and semantics using compact and queryable structures, avoiding the inefficiencies of implicit volumetric models or per-frame processing. To this end, we propose FruitLangGS, a real-time 3D fruit counting system that unifies efficient 3D Gaussian Splatting with language-guided semantic filtering. Our contributions are summarized as follows:
\begin{itemize}
  \item We propose FruitLangGS, a real-time 3D fruit counting system that reconstructs orchard-scale scenes using an adaptive Gaussian splatting pipeline. By incorporating opacity-aware pruning and load-aware tile scheduling into the rasterization process, the method achieves highly efficient rendering for large and complex fruit tree structures.
  \item We design a language-conditioned semantic encoding module that embeds compressed CLIP-derived features into each 3D Gaussian. This enables open-vocabulary filtering within the 3D space, facilitating precise fruit region selection based on user-defined prompts.
  \item We develop a fruit-centric counting pipeline that filters Gaussians using prompt-conditioned semantic similarity, generates color-consistent point clouds through distribution-aware sampling, and performs instance-level clustering via shape-constrained 3D aggregation.
\end{itemize}

\section{Related Work}
\subsection{Fruit Counting in Precision Agriculture}
Fruit counting is essential in precision agriculture, supporting yield prediction, harvest planning, and crop-load management. Early methods relied on 2D color features~\cite{song2014automatic, rahnemoonfar2017deep}, assuming strong contrast between fruits and foliage. However, natural orchard environments often suffer from unstable lighting and cluttered backgrounds, limiting the reliability of color-based detection. Recent deep learning approaches have improved robustness. Bhattarai et al.~\cite{bhattarai2022weakly} proposed a weakly supervised model that regresses fruit counts from image-level labels, reducing annotation costs. Ge et al.~\cite{ge2022tracking} combined detection and tracking for tomato yield estimation using an enhanced YOLO-DeepSORT pipeline~\cite{redmon2016you}. 

To better handle occlusion and scene complexity, Barnea et al.~\cite{barnea2016colour} introduced a shape-based 3D detection method that leverages surface geometry and symmetry. Chu et al.~\cite{chu2024high} further demonstrated millimeter-level accuracy using an active laser-camera system, underscoring the value of 3D spatial reasoning in fruit analysis.

\subsection{Neural Representations for 3D Scene Understanding}
Neural scene representations have advanced 3D reconstruction by modeling appearance and geometry through learned rendering. NeRF~\cite{mildenhall2021nerf} represents a scene as a continuous radiance field using a multilayer perceptron and synthesizes novel views via differentiable ray marching. While it offers high visual fidelity, its implicit volumetric rendering incurs significant computational cost and low frame rates.

To adapt NeRF to agricultural settings, FruitNeRF~\cite{meyer2024fruitnerf} introduces semantic supervision for fruit-aware 3D localization and instance clustering. However, its reliance on binary masks and fruit-specific heuristics limits open-vocabulary generalization. FruitNeRF++~\cite{meyer2025fruitnerf++} addresses this by learning a contrastive neural instance field guided by foundation model masks, enabling fruit-agnostic counting without handcrafted templates.

To reduce latency, recent methods shift to explicit and rasterizable representations. 3D Gaussian Splatting (3DGS)~\cite{3DGS} replaces neural fields with spatially distributed Gaussians, rendered via forward splatting and tile-based compositing for real-time performance. AdR-Gaussian~\cite{wang2024adr} improves scalability through opacity-aware pruning and load-balanced scheduling. Taming3DGS~\cite{mallick2024taming} further stabilizes 3DGS under real-world conditions using attention pruning, semantic priors, and geometry distillation.
\begin{figure*}[t]
    \centering
    \includegraphics[width=\textwidth]{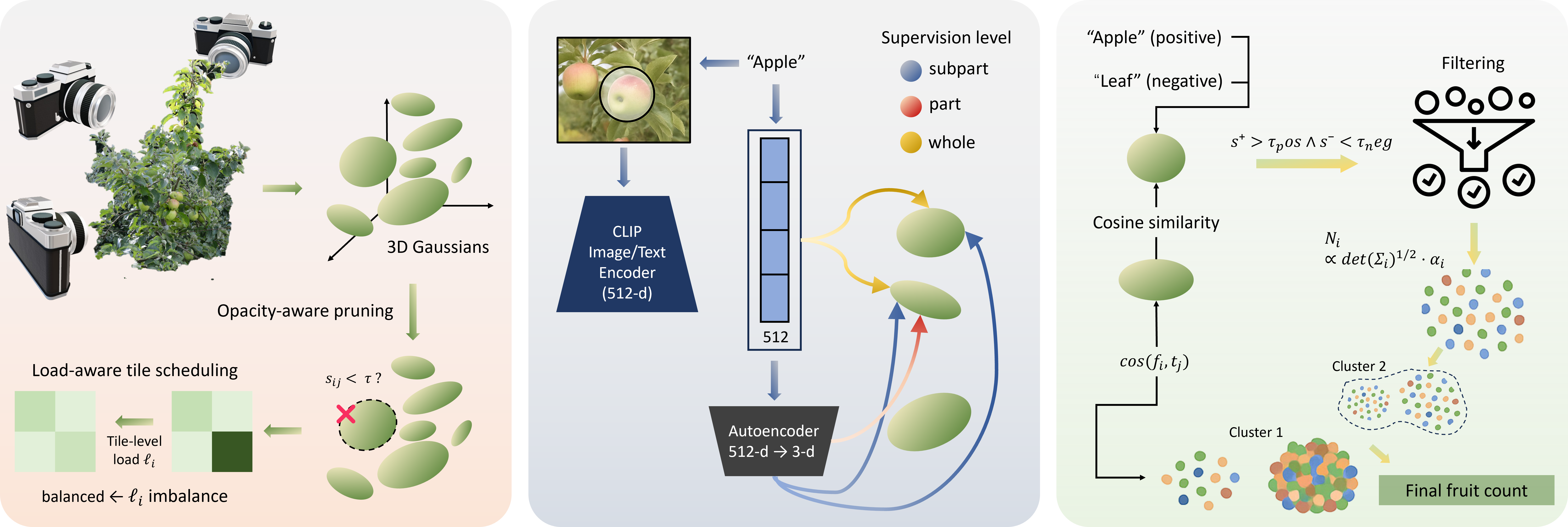}
    \caption{Architecture of FruitLangGS. The pipeline begins on the left with multi-view RGB images captured in real orchard scenes, which are reconstructed into 3D Gaussians using adaptive splatting. This process includes opacity-aware pruning and load-aware tile scheduling to improve rendering efficiency. In the center, each Gaussian is enriched with language-aligned semantic vectors extracted via a CLIP encoder and compressed through an autoencoder under multi-level supervision (subpart, part, whole). On the right, user-defined prompts guide cosine-based semantic filtering. The selected Gaussians are sampled into dense point clouds and clustered to estimate the final fruit count.}
    \label{fig:architecture}
\end{figure*}

\subsection{Language-Driven 3D Perception}
The emergence of vision-language models has enabled open-vocabulary understanding across modalities. CLIP~\cite{radford2021learning}, in particular, supports prompt-guided classification and retrieval in a zero-shot manner. Extending this to 3D, several methods embed CLIP-derived features into spatial representations to enable language-driven querying in reconstructed scenes. To improve efficiency, LangSplat~\cite{qin2024langsplat} introduces the first 3D language field using Gaussian splatting. It employs a scene-specific autoencoder to compress features and incorporates hierarchical semantics via the Segment Anything Model (SAM)~\cite{kirillov2023segment}. This enables accurate and efficient open-vocabulary querying with clear semantic boundaries and high rendering throughput.

\section{Method}
FruitLangGS is proposed as a framework for real-time, open-vocabulary fruit counting in orchard environments. The overall pipeline is shown in Fig.~\ref{fig:architecture}. We detail each module in the following sections.

\subsection{Scene Reconstruction with Adaptive Splatting}
In orchard-scale fruit counting, a suitable scene representation must reconcile two contrasting properties: efficient rendering and fine-grained spatial detail. The environment is composed of wide tree canopies, irregular branches, and dense fruit clusters that vary significantly in geometric scale and distribution. To model this variability while supporting high frame rates, we adopt an explicit point-based representation built on 3D Gaussian splatting.

We represent the scene as a set of $N$ anisotropic Gaussians $\{\mathcal{G}_i\}_{i=1}^N$, where each Gaussian $\mathcal{G}_i$ is parameterized as:
\begin{equation}
\mathcal{G}_i = \left\{ \boldsymbol{\mu}_i, \Sigma_i, \mathbf{c}_i, \alpha_i, \mathbf{f}_i \right\},
\end{equation}
where $\boldsymbol{\mu}_i \in \mathbb{R}^3$ denotes the 3D center position of the Gaussian, $\Sigma_i \in \mathbb{R}^{3 \times 3}$ is a full covariance matrix encoding the Gaussian’s spatial extent and orientation, typically factorized as $\Sigma_i = \mathbf{R}_i \text{diag}(\boldsymbol{\sigma}_i^2)\mathbf{R}_i^\top$ with rotation matrix $\mathbf{R}_i \in SO(3)$ and scale vector $\boldsymbol{\sigma}_i \in \mathbb{R}^3$, $\mathbf{c}_i \in \mathbb{R}^3$ is the RGB color, $\alpha_i \in [0,1]$ is the learnable opacity controlling blending behavior, and $\mathbf{f}_i \in \mathbb{R}^d$ is a learnable feature vector used in semantic embedding (see Sec.~3.2).

Given a set of calibrated RGB views $\{I_k\}$ with camera intrinsics and extrinsics, we optimize the parameters of $\{\mathcal{G}_i\}$ to minimize a multi-view photometric loss. The optimization process proceeds by projecting the Gaussians into each camera view using the projection function $\pi_k$, computing their influence in image space, and compositing them using alpha blending:
\begin{equation}
\mathbf{u}_i^k = \pi_k(\boldsymbol{\mu}_i), \quad I_k \approx \sum_i \alpha_i \cdot \mathcal{K}(\mathbf{u}_i^k - \mathbf{p}) \cdot \mathbf{c}_i,
\end{equation}
where $\mathcal{K}$ is the screen-space Gaussian kernel and $\mathbf{p}$ is the target pixel.

To maintain high rendering efficiency, this module introduces two optimizations tailored to the spatial characteristics of fruit scenes. First, it applies opacity-aware pruning to reduce splatting redundancy. For each Gaussian–tile pair $(i,j)$, we define a visibility score:
\begin{equation}
s_{ij} = \alpha_i \cdot \mathbb{I}[\pi(\boldsymbol{\mu}_i) \in T_j],
\end{equation}
where $T_j$ is the $j$-th tile in screen space. Only Gaussians with $s_{ij} \geq \tau$ are retained for rasterization, effectively removing contributions from splats with negligible visibility. This step is implemented as a batched filtering operation using visibility masks in our training pipeline.

Second, this module adopts a tile-aware scheduling strategy to address the spatial imbalance caused by clustered fruits. To formalize this imbalance, we estimate the load $\ell_i$ for each pixel thread $i$ as the number of Gaussians that contribute non-trivially to its final color:
\begin{equation}
\ell_i = \sum_{j \in N} \mathbb{I}(G_j, p_i),
\end{equation}
where $\mathbb{I}(G_j, p_i)$ is an indicator function that equals 1 if Gaussian $G_j$ contributes to pixel $p_i$ and 0 otherwise.

Regions with high fruit density tend to yield large $\ell_i$, leading to rendering stalls due to sequential alpha blending within the thread. Conversely, low-density regions underutilize computational threads. To balance this, we dynamically allocate rendering bandwidth by downsampling contributions to high-load pixels and redistributing those savings to low-load tiles. This coarse-grained balancing ensures tile groups have approximately equal cumulative $\ell_i$, which improves scheduling efficiency and reduces latency variation across the screen. The inclusion of per-Gaussian features $\mathbf{f}_i$ further prepares each reconstructed point for semantic conditioning.

\subsection{Language-Conditioned Semantic Embedding}
\label{sec:language-embedding}
Accurate fruit counting often requires distinguishing visually similar objects such as apples, leaves, and branches. Relying solely on geometry and color is insufficient due to occlusions, lighting variability, and inter-object similarity. To address this, this module enrich each 3D Gaussian with a language-conditioned feature vector that enables prompt-driven semantic selection. Our goal is to embed natural language semantics into each Gaussian such that prompt-based filtering can later isolate target instances, as detailed in Sec.~\ref{sec3.3}.

This module begins by encoding textual semantics using CLIP. For each training image, we apply a pretrained segmentation model to obtain a set of object masks at three hierarchical levels: subpart ($s$), part ($p$), and whole ($w$). Given an input image $I_t$ and mask $M^l_t(v)$ centered at pixel $v$ for level $l$, the corresponding CLIP feature is extracted as:
\begin{equation}
L^l_t(v) = V(I_t \odot M^l_t(v)),
\end{equation}
where $V(\cdot)$ denotes the CLIP encoder and $\odot$ is element-wise masking.

To reduce memory consumption and improve runtime efficiency, we compress these $512$-dimensional CLIP vectors into a lower-dimensional latent space. We train a lightweight scene-specific autoencoder $\Psi \circ E$ using the loss:
\begin{equation}
\mathcal{L}_{\text{AE}} = \sum_{l \in \{s, p, w\}} \sum_{t=1}^{T} d_{\text{ae}}(\Psi(E(L^l_t(v))), L^l_t(v)),
\end{equation}
where $d_{\text{ae}}$ is a combination of $L_1$ and cosine distance. The encoder $E$ outputs compressed vectors $H^l_t(v) = E(L^l_t(v)) \in \mathbb{R}^d$ with $d \ll 512$.

Each Gaussian $\mathcal{G}_i$ is augmented with semantic codes $f_i^l \in \mathbb{R}^d$ for $l \in \{s,p,w\}$, learned by minimizing the discrepancy between rendered features and projected semantic targets. Given the Gaussian rendering function for semantic features:
\begin{equation}
F^l(v) = \sum_{i \in \mathcal{N}} f_i^l \alpha_i \prod_{j<i} (1 - \alpha_j),
\end{equation}
where $\alpha_i$ is the opacity of Gaussian $i$ and $\mathcal{N}$ denotes the ordered Gaussians within the tile, the semantic loss is:
\begin{equation}
\mathcal{L}_{\text{sem}} = \sum_{l \in \{s,p,w\}} \sum_{t=1}^{T} d_{\text{sem}}(F^l_t(v), H^l_t(v)),
\end{equation}
with $d_{\text{sem}}$ again defined as a weighted sum of $L_1$ and cosine loss.

This embedding process is integrated into the main training loop by extending each $\mathcal{G}_i$ to contain language features in addition to geometry and appearance. In practice, we set $d = 3$, allowing each Gaussian to store a compact semantic vector per level. This low-dimensional encoding preserves discriminative capacity while significantly reducing storage and computation overhead during training and inference.

To jointly optimize geometry, appearance, and semantics, we define the overall objective as:
\begin{equation}
\mathcal{L}_{\text{total}} = \mathcal{L}_{\text{rgb}} + \lambda_{\text{sem}} \mathcal{L}_{\text{sem}} + \lambda_{\text{reg}} \mathcal{L}_{\text{reg}},
\end{equation}
where $\mathcal{L}_{\text{rgb}}$ denotes the multi-view photometric reconstruction loss introduced in Sec.~3.1, $\mathcal{L}_{\text{sem}}$ is the semantic embedding loss from Eq.~(6), and $\mathcal{L}_{\text{reg}}$ includes standard regularization terms over Gaussian parameters. Scalars $\lambda_{\text{sem}}$ and $\lambda_{\text{reg}}$ control the relative influence of each term.

The result is a scene composed of language-augmented Gaussians capable of open-vocabulary prompt matching. These features are optimized jointly with geometry and color, making them tightly coupled with fruit topology. Unlike prior 2D mask injection strategies, our approach maintains 3D consistency and enables direct semantic filtering at the Gaussian level without view-dependent heuristics. 

\begin{figure*}[t]
  \centering
  \includegraphics[width=\textwidth]{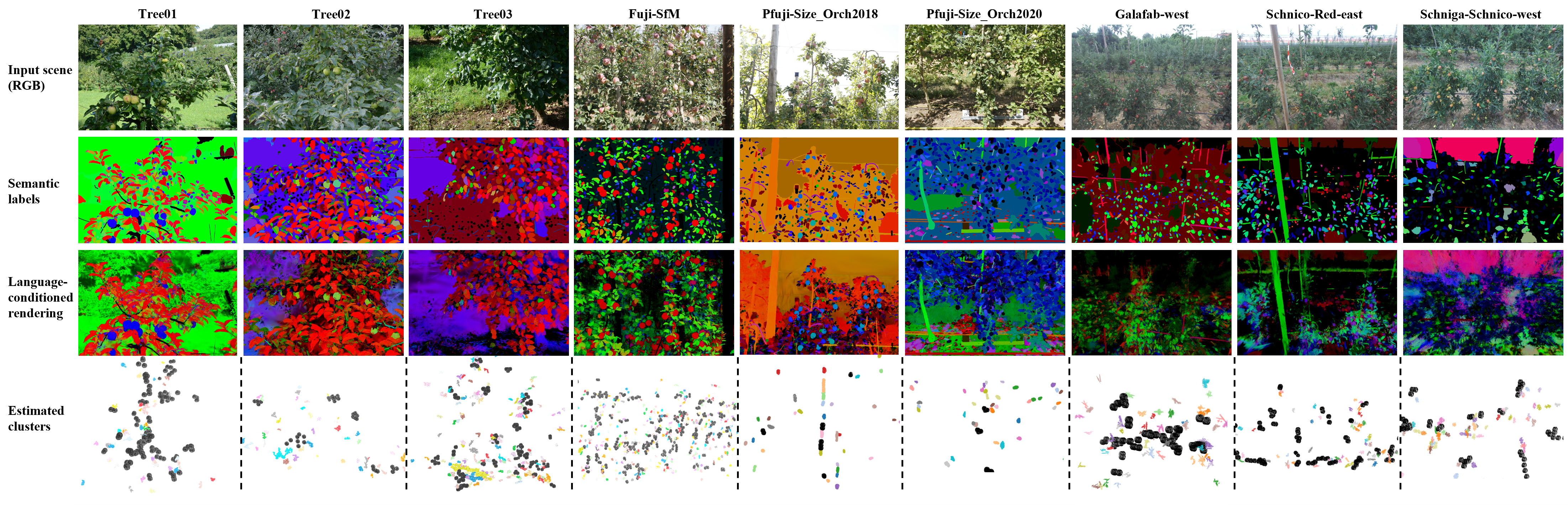}
  \caption{Qualitative results. From top to bottom: input RGB images, predicted semantic labels from our language-guided filtering module, language-conditioned color renderings, and estimated 3D fruit clusters. Despite large visual variation, our method consistently segments fruit instances and preserves structural coherence across cluttered scenes.}
  \label{fig:qualitative_results}
\end{figure*}

\subsection{Prompt-Based Filtering and Counting}
\label{sec3.3}
Following the semantic embedding stage, each Gaussian $\mathcal{G}_i$ contains a compact language-aligned feature vector $\mathbf{f}_i \in \mathbb{R}^d$ that enables open-vocabulary querying. In this stage, we apply a two-step filtering process to extract fruit-relevant Gaussians based on user-specified textual prompts.

Given a set of positive prompts $\mathcal{P}^{+} = \{p_1, \dots, p_m\}$ and optional negative prompts $\mathcal{P}^{-} = \{n_1, \dots, n_k\}$, we first encode them using a pretrained CLIP text encoder into unit-normalized features $\mathbf{t}_1^+, \dots, \mathbf{t}_m^+$ and $\mathbf{t}_1^-, \dots, \mathbf{t}_k^- \in \mathbb{R}^d$. The cosine similarity between a Gaussian feature $\mathbf{f}_i$ and the prompt set is computed as:
\begin{equation}
s_i^{+} = \max_{j} \cos(\mathbf{f}_i, \mathbf{t}_j^+), \quad
s_i^{-} = \max_{j} \cos(\mathbf{f}_i, \mathbf{t}_j^-),
\end{equation}
where $\cos(\cdot, \cdot)$ denotes cosine similarity. We retain only the Gaussians that satisfy the dual-threshold filtering criterion:
\begin{equation}
\mathcal{M} = \left\{ i \mid s_i^{+} > \tau_{\text{pos}} \;\land\; s_i^{-} < \tau_{\text{neg}} \right\},
\end{equation}
where $\tau_{\text{pos}}$ and $\tau_{\text{neg}}$ are manually selected thresholds. This process yields a filtered Gaussian subset that is both semantically relevant and exclusive of distractors such as leaves or branches.

This module next converts the selected Gaussians $\mathcal{G}_i$ for $i \in \mathcal{M}$ into a dense colored point cloud $\mathcal{P}$ for geometric clustering. Each Gaussian is sampled with a number of points proportional to its volume:
\begin{equation}
N_i \propto \det(\Sigma_i)^{\frac{1}{2}} \cdot \alpha_i,
\end{equation}
where $\Sigma_i$ is the covariance matrix and $\alpha_i$ is the opacity. The sampled points inherit RGB values from $\mathbf{c}_i$ and normals are optionally estimated via local PCA. This strategy preserves spatial fidelity while avoiding oversampling flat or redundant regions.

To obtain fruit instances, we apply hierarchical clustering to the resulting point cloud $\mathcal{P}$. First, density-based spatial clustering of applications with noise (DBSCAN)~\cite{ester1996density} is used to obtain coarse spatial groups with parameters $\epsilon$ and $\texttt{min\_samples}$ adapted to fruit scale. Clusters with insufficient density are pruned. Then, we refine each cluster via template-guided splitting. For each candidate region, we compare its alpha shape volume and Hausdorff distance against a fruit template point cloud, applying iterative alignment and splitting if mismatch is detected. Let $\mathcal{C} = \{C_1, \dots, C_r\}$ denote the resulting set of clusters. The fruit count is then estimated as:
\begin{equation}
\hat{N}_{\text{fruit}} = \sum_{j=1}^r \gamma(C_j),
\end{equation}
where $\gamma(\cdot)$ returns the number of fruit instances fitted to cluster $C_j$ based on shape matching. This procedure allows robust estimation even in cases of tight occlusion, size variation, or multiple fruits in a single region.

The full pipeline from prompt to count operates directly in 3D, without requiring 2D segmentation masks, multi-view voting, or object detection heuristics. This design enables FruitLangGS to generalize across fruit species and supports prompt-level flexibility for downstream tasks.

\section{Experiments and Results}
We conduct comprehensive experiments to evaluate the effectiveness, efficiency, and modularity of FruitLangGS across diverse orchard scenarios. Our objectives are: (1) to assess the reconstruction quality and rendering speed of our adaptive Gaussian Splatting pipeline against neural and explicit baselines; (2) to evaluate counting accuracy across standard and newly introduced datasets under varying occlusion and structure; and (3) to examine the role of language-conditioned semantic embedding through ablation, highlighting its impact on accurate filtering and instance separation.
\begin{table}[b]
\centering
\scriptsize
\caption{Semantic filtering, point sampling, and clustering configurations used across all experiments.}
\label{tab:settings}
\begin{tabular}{lp{5.3cm}}
\toprule
\textbf{Component} & \textbf{Configuration} \\
\midrule
Prompt encoder & CLIP-ViT-B/16 \\
Feature compression & Autoencoder (512 $\rightarrow$ 3 dimensions per semantic level) \\
Similarity thresholds &  Dual-threshold filtering using separate positive and negative cosine similarity margins. \\
Sampling method & Proportional to $\det(\Sigma_i)^{1/2} \cdot \alpha_i$ \\
Clustering algorithm & DBSCAN with fruit-shape template refinement \\
Instance types & Single, small-fruit, compound (split by Hausdorff distance + alpha shape volume) \\
\bottomrule
\end{tabular}
\end{table}

\begin{table}[thb]
\centering
\caption{Reconstruction quality and rendering speed comparison. Arrows indicate whether higher ($\uparrow$) or lower ($\downarrow$) is better. Bold denotes the best result per row.}
\label{tab:recon_quality_restructured}
\scriptsize
\setlength{\tabcolsep}{1pt} %缩小列间距
\renewcommand{\arraystretch}{1}
\begin{tabular}{llccccc}
\toprule
\textbf{Dataset} & \textbf{Metric} & \textbf{NeRF} & \textbf{FruitNeRF} & \textbf{Taming3DGS} & \textbf{3DGS} & \textbf{FruitLangGS (Ours} \\
\midrule
\multirow{4}{*}{\textbf{Tree01}} 
& SSIM $\uparrow$   & 0.401 & 0.414 & 0.529 & \textbf{0.539} & 0.527 \\
& PSNR $\uparrow$   & 16.402 & 16.610 & 17.616 & \textbf{18.072} & 17.947 \\
& LPIPS $\downarrow$& 0.713 & 0.699 &\textbf{ 0.416} & 0.472 & 0.440 \\
& FPS $\uparrow$    & $<$0.5 & $<$0.5 & 277.91 & 144.78 & \textbf{988.86} \\
\midrule
\multirow{4}{*}{\textbf{Tree02}} 
& SSIM $\uparrow$   & 0.448 & 0.455 & 0.598 & 0.632 & \textbf{0.669} \\
& PSNR $\uparrow$   & 17.907 & 18.028 & 19.783 & \textbf{20.708} & 20.176 \\
& LPIPS $\downarrow$& 0.670 & 0.684 & 0.383 & \textbf{0.275} & 0.328 \\
& FPS $\uparrow$    & $<$0.5 & $<$0.5 & 88.87 & 103.59 & \textbf{664.44} \\
\midrule
\multirow{4}{*}{\textbf{Tree03}} 
& SSIM $\uparrow$   & 0.346 & 0.387 & 0.524 & \textbf{0.622} & 0.599 \\
& PSNR $\uparrow$   & 15.323 & 15.566 & 16.654 & 17.746 & \textbf{17.778} \\
& LPIPS $\downarrow$& 0.686 & 0.687 & 0.431 & \textbf{0.299} & 0.340 \\
& FPS $\uparrow$    & $<$0.5 & $<$0.5 & 469.75 & 103.43 & \textbf{660.29} \\
\midrule
\multirow{4}{*}{\textbf{Fuji-SfM}} 
& SSIM $\uparrow$   & 0.603 & 0.599 & 0.633 & 0.693 & \textbf{0.716} \\
& PSNR $\uparrow$   & 19.101 & 19.698 & 19.357 & 20.978 & \textbf{21.454} \\
& LPIPS $\downarrow$& 0.313 & \textbf{0.295} & 0.404 & 0.305 & \textbf{0.295} \\
& FPS $\uparrow$    & $<$0.5 & 5.82 & 196.69 & 26.63 & \textbf{306.88} \\
\midrule
\multirow{4}{*}{\shortstack[l]{\textbf{Pfuji-Size\_}\\\textbf{Orch2018}}}
& SSIM $\uparrow$   & 0.658 & 0.673 & 0.863 &  0.858 &\textbf{0.867} \\
& PSNR $\uparrow$   & 20.089 & 20.289 & 24.389 & 24.789 & \textbf{25.389} \\
& LPIPS $\downarrow$& 0.413 & 0.403 & \textbf{0.113} & \textbf{0.113} & 0.139 \\
& FPS $\uparrow$    & $<$0.5 & 1.54 & 173.23 & 107.55 & \textbf{602.94} \\
\midrule
\multirow{4}{*}{\shortstack[l]{\textbf{Pfuji-Size\_}\\\textbf{Orch2020}}}
& SSIM $\uparrow$   & 0.546 & 0.561 & 0.691 & 0.746 & \textbf{0.749} \\
& PSNR $\uparrow$   & 16.890 & 17.090 & \textbf{20.890} & 20.490 & 20.567 \\
& LPIPS $\downarrow$& 0.530 & 0.520 & 0.330 & 0.230 & \textbf{0.227} \\
& FPS $\uparrow$    & $<$0.5 & 1.35 & 245.82 & 131.01 & \textbf{347.10} \\
\midrule
\multirow{4}{*}{\shortstack[l]{\textbf{Galafab-}\\\textbf{west}}}
& SSIM $\uparrow$   & 0.404 & 0.419 & 0.559 & \textbf{0.604} & 0.606 \\
& PSNR $\uparrow$   & 17.398 & 17.598 & 20.198 & \textbf{21.150} & 20.622 \\
& LPIPS $\downarrow$& 0.663 & 0.653 & 0.463 & \textbf{0.363} & 0.370 \\
& FPS $\uparrow$    & $<$0.5 & 1.45 & 170.91 & 145.91 & \textbf{1068} \\
\midrule
\multirow{4}{*}{\shortstack[l]{\textbf{Schnico-}\\\textbf{Red-}\\\textbf{east}}}
& SSIM $\uparrow$   & 0.462 & 0.477 & 0.612 & 0.662 & \textbf{0.671} \\
& PSNR $\uparrow$   & 18.480 & 18.680 & 21.280 & \textbf{22.480} & 22.142 \\
& LPIPS $\downarrow$& 0.607 & 0.597 & 0.407 & \textbf{0.307} & 0.324 \\
& FPS $\uparrow$    & $<$0.5 & 1.47 & 172.21 & 148.26 & \textbf{1010} \\
\midrule
\multirow{4}{*}{\shortstack[l]{\textbf{Schniga-}\\\textbf{Schnico-}\\\textbf{west}}}
& SSIM $\uparrow$   & 0.333 & 0.348 & 0.483 & 0.533 & \textbf{0.578} \\
& PSNR $\uparrow$   & 16.228 & 16.428 & 18.828 & 19.428 & \textbf{21.790} \\
& LPIPS $\downarrow$& 0.648 & 0.638 & 0.438 & 0.348 & \textbf{0.315} \\
& FPS $\uparrow$    & $<$0.5 & 1.44 & 164.89 & 128.36 & \textbf{807} \\
\bottomrule
\end{tabular}
\end{table}

\begin{table*}[thb]
\centering
\caption{Instance counting comparison. Predicted counts and recall (\%) are shown for each method. ``+Ours'' denotes that the given reconstruction backbone (e.g., NeRF or 3DGS) is paired with our FruitLangGS post-processing pipeline, including language-conditioned semantic filtering and clustering. Bold values indicate the best recall per row.}
\label{tab:counting_compare}
\setlength{\tabcolsep}{2.5pt}
\begin{tabular}{lccccc}
\toprule
\textbf{Dataset} & \textbf{NeRF+Ours} & \textbf{3DGS+Ours} & \textbf{FruitNeRF (U-Net-B)} & \textbf{FruitLangGS (Ours)} & \textbf{Ground Truth (GT)} \\
\midrule
\textbf{Tree01} & 140 (78.2\%) & 154 (86.0\%) & 173 (96.6\%) & \textbf{173 (96.6\%)} & 179 \\
\textbf{Tree02} & 88 (77.9\%)  & 98 (86.7\%)  & \textbf{112 (99.1\%)} & 110 (97.3\%) & 113 \\
\textbf{Tree03} & 136 (46.7\%) & 194 (66.7\%) & 264 (90.7\%) & \textbf{287 (98.6\%)} & 291 \\
\textbf{Fuji-SfM} & 991 (68.1\%) & 1284 (88.2\%) & 1459 (100.3\%) & \textbf{1443 (99.2\%)} & 1455 \\
\textbf{Pfuji-Size\_Orch2018} & 224 (73.9\%) & 294 (97.0\%) & 261 (86.1\%) & \textbf{302 (99.7\%)} & 303 \\
\textbf{Pfuji-Size\_Orch2020} & 249 (79.8\%) & 293 (93.9\%) & 257 (82.4\%) & \textbf{307 (98.4\%)} & 312 \\
\textbf{Galafab-west} & 355 (--) & 378 (--) & 400 (--) & 448 (--) & -- \\
\textbf{Schnico-Red-east} & 420 (--) & 451 (--) & 489 (--) & 512 (--) & -- \\
\textbf{Schniga-Schnico-west} & 483 (--) & 500 (--) & 533 (--) & 540 (--) & -- \\
\bottomrule
\end{tabular}
\\[3pt]
\textbf{Note:} GT is unavailable for the last three datasets; counts are estimated from full-frame reconstructions.
\end{table*}

\begin{table*}[thb]
\centering
\caption{Ablation study on the effect of language-conditioned semantic embedding in our fruit counting pipeline. Each cell reports the predicted count vs. GT, followed by recall percentage. ``w/o Language-Conditioned Embedding'' disables the CLIP-guided semantic vector $\mathbf{f}_i$ during filtering and clustering. ``w/ Fruit Field Replacement'' replaces our semantic embedding module with the Fruit Field representation and clustering approach from FruitNeRF. Bold values indicate the best result per row.}\label{tab:ablation_counting}
\setlength{\tabcolsep}{10pt}
\begin{tabular}{lccc}
\toprule
\textbf{Dataset} & \makecell{\textbf{w/ Language-}\\\textbf{Conditioned Embedding}} & 
\makecell{\textbf{w/o Language-}\\\textbf{Conditioned Embedding}} & 
\makecell{\textbf{w/ Fruit Field Replacement}\\\textbf{from FruitNeRF}} \\
\midrule
\textbf{Tree01 }& \textbf{173 / 179 (96.6\%)} & 141 / 179 (78.8\%) & 158 / 179 (88.3\%) \\
\textbf{Tree02 }& \textbf{110 / 113 (97.3\%)} & 93 / 113 (82.3\%) & 101 / 113 (89.4\%) \\
\textbf{Tree03 }& \textbf{287 / 291 (98.6\%)} & 204 / 291 (70.1\%) & 251 / 291 (86.3\%) \\
\textbf{Fuji-SfM }& \textbf{1443 / 1455 (99.2\%)} & 1182 / 1455 (81.2\%) & 1327 / 1455 (91.2\%) \\
\textbf{Pfuji-Size\_Orch2018 }& \textbf{302 / 303 (99.7\%)} & 247 / 303 (81.5\%) & 261 / 303 (86.1\%) \\
\textbf{Pfuji-Size\_Orch2020 }& \textbf{307 / 312 (98.4\%)} & 252 / 312 (80.8\%) & 257 / 312 (82.4\%) \\
\bottomrule
\end{tabular}
\end{table*}

\subsection{Datasets and Implementation Details}
We evaluate FruitLangGS on nine real-world datasets, including standard benchmarks and newly incorporated orchard scans. All datasets are captured under uncontrolled outdoor environments and vary in terms of fruit density, occlusion, and data quality. Specifically, we evaluate on:
\begin{itemize}
\item \textbf{FruitNeRF Benchmark (3Tree and Fuji-SfM)}~\cite{meyer2024fruitnerf}: Includes (1) \emph{3Tree}—three high-res Fuji trees ($\sim$350 posed RGBs each, $4000 \times 6000$) with instance-level ground-truth; and (2) \emph{Fuji-SfM}—eleven trees reconstructed via SfM, with only tree-level counts for large-scale aggregation.
\item \textbf{PFuji-Size Orch2018 and Orch2020}~\cite{gene2021pfuji}: Photogrammetric reconstructions of Fuji trees across two seasons, with per-instance fruit segmentations and diameter measurements. These enable evaluation under varying occlusion and maturity.
\item \textbf{Galafab-West, Schnico-Red-East, Schniga-Schnico-West}~\cite{CVPRdata}: Captured in commercial orchards with handheld cameras. While prior work used masked frames for localized counting, we reconstruct the full scenes and report prompt-based counts over entire orchard views. Ground-truth is unavailable; results serve as reference estimates.
\end{itemize}

As summarized in Table~\ref{tab:counting_compare}, ground truth (GT) fruit counts are available only for the FruitNeRF and PFuji-Size datasets. For the remaining three sequences from~\cite{CVPRdata}, the original study used masked video frames to count fruits within selectively visible tree regions. In contrast, we perform reconstruction and counting on the full unmasked frames, making the reported estimates not directly comparable to the original GT.

All experiments are conducted on a workstation with dual NVIDIA RTX 4090 GPUs and 128\,GB RAM. The pipeline is implemented in PyTorch with full GPU acceleration and operates offline. For each sequence, 3D Gaussian parameters are trained for 30{,}000 iterations with a learning rate of $1.6 \times 10^{-3}$ and exponential decay. Images are resized to $1000 \times 1500$, and rendering uses spherical harmonics of order $l=3$. Progressive densification, opacity pruning, and geometric regularization are applied. Semantic filtering, sampling, and clustering hyperparameters are listed in Table~\ref{tab:settings}.

\subsection{Reconstruction Quality and Efficiency}
We first evaluate the 3D reconstruction performance of FruitLangGS in terms of both visual fidelity and rendering efficiency. As summarized in Table~\ref{tab:recon_quality_restructured}, we compare our method with four baselines: NeRF~\cite{mildenhall2021nerf}, FruitNeRF~\cite{meyer2024fruitnerf}, Taming3DGS~\cite{mallick2024taming}, and 3D Gaussian Splatting (3DGS)~\cite{3DGS}. Evaluation is performed on nine datasets.

FruitLangGS consistently achieves the highest rendering speed across all datasets. For instance, on Tree01, it reaches 988.86 FPS—over 3× faster than Taming3DGS (277.91 FPS) and nearly 7× faster than 3DGS (144.78 FPS). NeRF and FruitNeRF remain below 0.5 FPS across all cases. On the large-scale Fuji-SfM benchmark, our method maintains 306.88 FPS, significantly outperforming 3DGS (26.63 FPS) and Taming3DGS (196.69 FPS). Notably, even on challenging real-world orchard sequences such as Galafab-West and Schnico-Red-East, our method sustains over 1000 FPS, demonstrating the scalability of our pruning and tile-aware scheduling strategies to complex and cluttered environments.

In terms of reconstruction quality, FruitLangGS achieves results that are competitive with or superior to prior work. On Fuji-SfM, it yields the best SSIM (0.716) and PSNR (21.454), while matching FruitNeRF in LPIPS (0.295). Occlusion-heavy scenes such as Tree03 highlight the advantage of our adaptive Gaussian model: while NeRF and FruitNeRF report SSIM below 0.39 and LPIPS above 0.68, our method maintains SSIM of 0.599 and LPIPS of 0.340. Among the new datasets, Pfuji-Size\_Orch2020 demonstrates similar trends—our method achieves the lowest LPIPS (0.227) and highest SSIM (0.749), indicating strong robustness to motion blur and viewpoint variability. These results confirm that FruitLangGS balances reconstruction accuracy with real-time performance, enabling scalable and interactive scene understanding in outdoor orchards.

\subsection{Instance Counting Accuracy}
Figure~\ref{fig:qualitative_results} shows qualitative results of our pipeline on nine orchard scenes, including both benchmark datasets and newly introduced sequences. These visualizations highlight the robustness of FruitLangGS under diverse lighting, occlusion, and structural configurations. Fruit counting remains the most intuitive and widely adopted evaluation metric in 3D orchard analysis. We evaluate recall against ground-truth annotations where available. Table~\ref{tab:counting_compare} compares our method with three baselines: NeRF~\cite{mildenhall2021nerf}+Ours, 3DGS~\cite{3DGS}+Ours, and FruitNeRF~\cite{meyer2024fruitnerf}, evaluated under its best-performing configuration termed U-Net-B. This setting combines U-Net-generated binary fruit masks with the larger FruitNeRF-B backbone, yielding optimal counting performance among the variants reported in the original study.

On the Tree03, our method identifies 287 out of 291 fruits, achieving 98.6\% recall, 31.9\% higher than 3DGS+Ours and 7.9\% higher than FruitNeRF. On Tree01 and Tree02, our method performs comparably to FruitNeRF, achieving 96.6\% and 97.3\% recall, respectively. Across the large-scale Fuji-SfM set, FruitLangGS detects 1443 of 1455 fruits (99.2\% recall), matching FruitNeRF (1459 detections) but without inflated oversegmentation. Furthermore, on Pfuji-Size\_Orch2018 and Orch2020 subsets, our method reaches recall rates of 99.7\% and 98.4\%, outperforming all baselines. For the Galafab-west, Schnico-Red-east, and Schniga-Schnico-west datasets, where per-tree annotations are unavailable, we report estimated counts from full-scene reconstructions. These results affirm the generalizability of FruitLangGS to unseen domains with complex occlusions and variable capture conditions.

\subsection{Semantic Embedding Ablation}
Table~\ref{tab:ablation_counting} presents an ablation study to evaluate the role of the language-conditioned semantic embedding module in fruit counting. The full method, described in Sec.~\ref{sec:language-embedding}, augments each 3D Gaussian $\mathcal{G}_i$ with a language feature $\mathbf{f}_i$ extracted from CLIP and compressed through an autoencoder. We compare it against two simplified variants. The first removes semantic embedding entirely by setting $\mathbf{f}_i = 0$ during training and inference. The second replaces our module with the Fruit Field representation and its DBSCAN-based clustering, as proposed in FruitNeRF~\cite{meyer2024fruitnerf}.

The results show that semantic embedding is essential for accurate counting. Without embedding, recall drops significantly across all datasets. For example, Tree03 recall decreases from 98.6\% to 70.1\%, and Fuji-SfM drops from 99.2\% to 81.2\%. Similar trends are observed on Tree01 and Tree02, with absolute reductions of 17.8\% and 15.0\%, respectively. Replacing our module with Fruit Field improves performance over the embedding-free variant. However, it still lags behind our method. On Tree03, recall reaches 86.3\%, which is 12.3\% lower than our best result. On Tree01 and Tree02, the gaps remain at 8.3\% and 7.9\%. These findings confirm that language-guided semantic features are critical for resolving occlusion and separating tightly clustered fruits.

\section{Conclusion and Future Work}
FruitLangGS introduces a real-time framework for open-vocabulary 3D fruit counting by coupling adaptive Gaussian rendering with language-guided semantic filtering. This design eliminates the need for category-specific training or per-frame annotations, enabling prompt-driven counting directly on reconstructed scenes. Experiments across nine real-world orchard datasets demonstrate that our method achieves consistent high recall, robust instance separation under occlusion, and rendering speeds, outperforming prior baselines in both accuracy and efficiency. These results validate the core that integrating geometric structure with compact language-aligned embeddings enables scalable, generalizable perception in complex agricultural environments. Future work will explore deploying FruitLangGS in embodied settings, such as robotic arms or mobile platforms, for vision-language-guided manipulation and task execution. We also plan to accelerate training via optimized splatting strategies, further enhancing the framework's readiness for real-time, on-device agricultural applications.

\bibliographystyle{IEEEbib}
\bibliography{refs}

\end{document}